\documentclass{article}

\usepackage[preprint, nonatbib]{neurips_2024}
\usepackage{subfigure}
\usepackage{subcaption}
\usepackage{amsmath} 
\usepackage[ruled,lined]{algorithm2e}
\usepackage[utf8]{inputenc} % allow utf-8 input
\usepackage[T1]{fontenc}    % use 8-bit T1 fonts
\usepackage{hyperref}       % hyperlinks
\usepackage{url}            % simple URL typesetting
\usepackage{booktabs}       % professional-quality tables
\usepackage{amsfonts}       % blackboard math symbols
\usepackage{nicefrac}       % compact symbols for 1/2, etc.
\usepackage{microtype}      % microtypography
\usepackage{xcolor}         % colors

\usepackage{graphicx} 
\usepackage[normalem]{ulem}     %%%% to insert the underline in table 
\useunder{\uline}{\ul}{}        %%%% to insert the underline in table
\usepackage{wrapfig}

\usepackage{xcolor}
\definecolor{mypink}{rgb}{0.95, 0.5, 0.7}
\definecolor{darkbrown}{rgb}{0.6, 0.2, 0.2}     
\definecolor{lightbrown}{rgb}{0.9, 0.3, 0.3}   
\definecolor{darkblue}{rgb}{0.0, 0.2, 0.8}     
\definecolor{lightblue}{rgb}{0.0, 0.5, 0.9}     
\hypersetup{
    colorlinks=true,
    linkcolor=green,
    filecolor=magenta,      
    urlcolor=mypink,
    citecolor=cyan
}    
\usepackage{fancyhdr}

\title{FANFOLD: Graph Normalizing Flows-driven Asymmetric Network for Unsupervised Graph-Level Anomaly Detection}

\author{%
Rui Cao$^{1 \footnotemark[2]}$ \quad Shijie Xue$^{1\footnotemark[2]}$ \quad Jindong Li$^{1\footnotemark[2]}$ \quad Qi Wang$^{2\footnotemark[1]}$ \quad Yi Chang$^2$ \\
\\
$^1$School of Artificial Intelligence, Jilin University \\
\quad $^2$Engineering Research Center of Knowledge-Driven Human-Machine Intelligence,\\
Ministry of Education, China\\ \quad
\texttt{\{caorui22, sjxue21, jdli21\}@mails.jlu.edu.cn}\\
\texttt{\{qiwang, yichang\}@jlu.edu.cn}\\
}
\begin{document}

\maketitle
\renewcommand{\thefootnote}{\fnsymbol{footnote}}
%Based on the widely acknowledged fact that anomalies tend to occur in low-density regions of a distribution. 

\begin{abstract}
Unsupervised graph-level anomaly detection (UGAD) has attracted increasing interest due to its widespread application. In recent studies, knowledge distillation-based methods have been widely used in unsupervised anomaly detection to improve model efficiency and generalization. However, the inherent symmetry between the source (teacher) and target (student) networks typically results in consistent outputs across both architectures, making it difficult to distinguish abnormal graphs from normal graphs. Also, existing methods mainly rely on graph features to distinguish anomalies, which may be unstable with complex and diverse data and fail to capture the essence that differentiates normal graphs from abnormal ones. In this work, we propose a Graph Normalizing \textbf{F}lows-driven \textbf{A}symmetric \textbf{N}etwork \textbf{FO}r Unsupervised Graph-\textbf{L}evel Anomaly \textbf{D}etection (FANFOLD in short). We introduce normalizing flows to unsupervised graph-level anomaly detection due to their successful application and superior quality in learning the underlying distribution of samples. Specifically, we adopt the knowledge distillation technique and apply normalizing flows on the source network, achieving the asymmetric network. In the training stage, FANFOLD transforms the original distribution of normal graphs to a standard normal distribution. During inference, FANFOLD computes the anomaly score using the source-target loss to discriminate between normal and anomalous graphs. We conduct extensive experiments on 15 datasets of different fields with 9  baseline methods to validate the superiority of FANFOLD. The code is available at \url{https://github.com/Goldenhorns/FANFOLD}.\footnotetext[2]{Equal contributions}\footnotetext[1]{Corresponding author}
\end{abstract}

\renewcommand{\thefootnote}{\fnsymbol{footnote}}

\section{Introduction}
%Graph data, found ubiquitously in various domains such as social networks, biology, and transportation systems, consists of nodes representing entities and edges denoting connections between them, showcasing its widespread applicability \cite{2020_arXiv_TuDataset}. 

Graph-level anomaly detection aims to identify anomalous patterns or outliers within the overall structure or behavior of graphs \cite{2021_TKDE_Survey_GAD_maxiaoxiao, 2022_WSDM_GLocalKD, 2023(2024)_NeurIPS_SIGNET}, which plays a crucial role in detecting potential threats across various domains such as social networks, biological systems, and transportation networks \cite{2020_arXiv_TuDataset}. Unsupervised graph-level anomaly detection utilizes algorithms to automatically identify irregularities within graphs, providing a scalable and adaptable solution that detects new anomalies without requiring labeled data. Recently, knowledge distillation (KD)-based approaches have been widely adopted in unsupervised anomaly detection tasks due to their unique ability to enhance model efficiency and generalization capabilities \cite{2020_CVPR_Uninformed_Students, 2022_IJCV_GCAD, 2021_CVPR_Anomaly_Detection, 2021_arXiv_Student_Teacher_feature_pyramid, 2021_Entropy_DTSNE}. The rationale behind this approach lies in the fact that the target network, having been exclusively trained on normal data, can only effectively replicate the source network's outputs on such data. Subsequently, during the testing phase, the disparity between the outputs of the target network and source network serves as an indicator of anomalies. Despite several attempts that have been made to apply knowledge distillation for unsupervised graph-level anomaly detection \cite{2023_WSDM_GOOD-D, 2023_DASFAA_TUAF, 2023_ECMLPKDD_CVTGAD, 2023_ECMLPKDD_HimNet}, there still exist certain issues.
%Unsupervised graph-level anomaly detection leverages algorithms to automatically uncover irregularities or outliers within a graph's structure or behavior, eliminating the dependency on labeled data for training. Such approach offers scalability across large datasets, adaptability to evolving patterns, and the ability to detect previously unseen anomalies without the need for manual labeling, thus facilitating more comprehensive and efficient anomaly detection in complex systems \cite{2023_WSDM_GOOD-D, 2023_DASFAA_TUAF, 2023_ECMLPKDD_CVTGAD, 2023_ECMLPKDD_HimNet}.

%Initially, the source network undergoes training on a pretext task aimed at acquiring a semantic embedding. Following this, the target network is then trained to replicate the output of the source network. The rationale behind this approach lies in the fact that the target network, having been exclusively trained on normal data, can only effectively replicate the source network's outputs on such data. Subsequently, during testing, the disparity between the outputs of the target network and source network serves as an indicator of anomalies. It is posited that this discrepancy is more pronounced for anomalous instances compared to their defect-free counterparts \cite{2022_WSDM_GLocalKD, 2022_ScientificReports_GLADC, 2023_WACV_AST}.

%--------------------- Fig 1----------------------%
\begin{figure}[!t]
    \centering
    \includegraphics[width=0.99\textwidth]{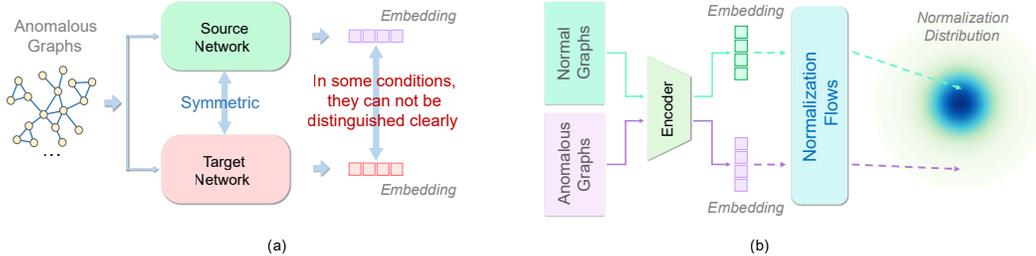}
    \caption{Toy examples of (a) the challenge caused by symmetric architecture and (b) the mapping of normalizing flows for both normal graphs and anomalous graphs.} 
    \label{fig: fig_1}
    % \vspace{-2em}
\end{figure}

%But, symmetric network may not always effectively differentiate between anomalous and normal graphs, leading to suboptimal performance in graph-level anomaly detection. 

%when knowledge distillation based models are employed in anomaly detection tasks, the process commences with the training of the teacher network using normal graphs, followed by the student network learning the node embeddings for normal graphs as delineated by the teacher network. In the testing phase, those graphs for which the embeddings exhibit a substantial deviation between the student and teacher networks are considered anomalous. 
\textbf{Firstly}, existing knowledge distillation based graph-level anomaly detection models mainly adopt symmetric networks, indicating that the target network (student) and source network (teacher) have similar or even identical architectures, e.g., existing works mainly adopt two identical GNN encoders for the source network and target network in their framework \cite{2022_WSDM_GLocalKD, 2022_ScientificReports_GLADC, 2023_ADMA_GLADST}. 
Nevertheless, in graph-level anomaly detection tasks, the challenge arises from the fact that traditional knowledge distillation approaches expect larger discrepancies between the outputs of two networks for detecting potential anomalies. 
However, as shown in Figure \ref{fig: fig_1} (a), the inherent symmetry between the source (teacher) network and target (student) network typically yields consistent outputs across both architectures, thereby making it difficult to distinguish abnormal graphs from normal graphs.

\textbf{Secondly}, traditional knowledge distillation based anomaly detection methods mainly involve a simple architecture, e.g., an encoder, with the main goal of improving the encoder's feature embedding ability. However, these methods may demonstrate instability when processing complex and diverse graph data. In fact, these obtained embeddings can be distinguished from the perspective of their distribution in the latent space \cite{2022_WSDM_GLocalKD, 2023_WSDM_GOOD-D, 2023_ECMLPKDD_CVTGAD}, which can reduce the perturbation caused by specific features on detection. 
More importantly, adopting a distribution-centric approach enables a more accurate capture of the intrinsic characteristics of normal graphs (e.g., as shown in Figure \ref{fig: fig_1} (b)), which in turn augments the sensitivity and accuracy of anomaly detection algorithms. Therefore, there is an urgent need for a method that can consider anomaly detection from the perspective of sample embedding distribution and distribution density.

%In fact, when obtaining an embedding, these embeddings can be distinguished from the perspective of distribution in the latent space \cite{2022_WSDM_GLocalKD, 2023_WSDM_GOOD-D, 2023_ECMLPKDD_CVTGAD}, which is very important for graph-level anomaly detection. 

To tackle the aforementioned challenges, we propose a novel graph normalizing flows-driven asymmetric network for unsupervised graph-level anomaly detection, named FANFOLD. Specifically, based on the widely acknowledged fact that anomalies tend to occur in low-density areas of a distribution \cite{2022_ICLR_GANF}, we introduce normalizing flows to map the empirical distribution of typical graph structures onto a Gaussian standard distribution within the source network. This mapping is designed to consider both the distribution of the sample embeddings and their density, thereby establishing the desired asymmetry within the network architecture. In the training stage, we first train an encoder via a reconstruction task on the source network, then we further train the normalizing flows based on the pre-trained encoder to achieve transformation from the original distribution of normal graphs to a standard normal distribution. In the inference stage, we utilize the difference in output between the source network and the target network to discriminate normal graphs and anomalous graphs. Our contributions are summarized as follows:
\begin{itemize}
    \item  We propose a novel graph normalizing flows-driven asymmetric network for unsupervised graph-level anomaly detection (FANFOLD in sort). To our best knowledge, we are the first to apply the graph normalizing flows and asymmetric network to unsupervised graph-level anomaly detection tasks, integrating their advantages and enabling them to work together efficiently.
    %We adopt the knowledge distillation technique and we apply normalizing flows on the source network, achieving the asymmetric network.
    \item We present an innovative use of a pretrained encoder and normalizing flows to transform the distribution of normal graph data into a standard Gaussian distribution. Such a method considers anomaly detection from the perspective of sample distribution, enabling the capture of more fundamental differences between normal and abnormal graphs.
    %coupled with a novel inference technique that leverages output discrepancies between source and target networks to accurately distinguish between normal and anomalous graphs.
    \item We conduct extensive experiments against 15 datasets from various fields with 9 representative baselines to validate FANFOLD's superiority on unsupervised graph-level anomaly detection tasks.
\end{itemize}

%In the training phase, we initially train an encoder through a reconstruction task on the source network, thereby establishing a foundational representation. Subsequently, we refine the normalizing flows based on this pretrained encoder, facilitating the transformation from the original distribution of normal graphs to a standard Gaussian distribution. During the inference phase, we exploit the disparity in the outputs of the source and target networks as a discriminative criterion to differentiate between normal and anomalous graphs.

\section{Related Work}
\subsection{Graph-level Anomaly Detection}
Graph-level anomaly detection aims to identify abnormal graphs among normal ones. These anomalous graphs often represent a minority but contain crucial patterns \cite{2021_TKDE_Survey_GAD_maxiaoxiao}. In recent years, there has been a surge of noteworthy scholarly endeavors. 
OCGIN \cite{2021_BigData_OCGIN} is the first representative model that integrates one-class classification with the graph isomorphism network (GIN) \cite{2019_ICLR_GIN}. This integration significantly enhances the capability and accuracy of graph-level anomaly detection by combining the strengths of both techniques. 
% OCGTL \cite{2022_IJCAI_OCGTL} integrates the strengths of deep one-class classification and neural transformation learning. 
GLocalKD \cite{2022_WSDM_GLocalKD} implements joint random distillation to detect both local and global graph anomalies. This is achieved by training one graph neural network to predict the output of another graph neural network, which has its weights fixed at random initialization.
GOOD-D \cite{2023_WSDM_GOOD-D} presents a novel perturbation-free graph data augmentation that avoids introducing perturbations. It employs hierarchical contrastive learning to enhance the detection of anomalous graphs by identifying semantic inconsistencies at multiple levels, thus improving the overall accuracy and robustness of anomaly detection in graph data.
% TUAF \cite{2023_DASFAA_TUAF} builds triple-unit graphs and further learns triple representations to simultaneously capture abundant information on edges and their corresponding nodes. 
% CVTGAD \cite{2023_ECMLPKDD_CVTGAD} applies transformer and cross-attention into UGAD, directly exploiting relationships across different views. 
SIGNET \cite{2023(2024)_NeurIPS_SIGNET} presents a multi-view subgraph information bottleneck framework. It derives anomaly scores and delivers explanations at the subgraph level.

\subsection{Normalizing Flows}
Normalizing flows are a family of generative models that create tractable distributions, allowing for efficient and exact sampling and density evaluation. A normalizing flow is a transformation of a complex distribution into a more simple probability distribution (e.g., a standard normal) by a sequence of mappings that are usually invertible and differentiable \cite{2020_TPAMI_Survey_Normalizing_Flows}.
RealNVP \cite{2017_ICLR_RealNVP} expands the scope of these models by incorporating real-valued non-volume preserving (real NVP) transformations. These transformations are powerful, stably invertible, and learnable, leading to an unsupervised learning algorithm that offers exact log-likelihood computation, precise sampling, efficient inference of latent variables, and an interpretable latent space.
DifferNet \cite{2021_WACV_DifferNet} utilizes a multi-scale feature extractor that allows the normalizing flow to assign meaningful likelihoods to the images. As normalizing flows are well-suited to handle low-dimensional data distributions, it leverages the descriptive features extracted by convolutional neural networks to estimate their density via normalizing flows.
AST \cite{2023_WACV_AST} focuses on industrial defect detection tasks (which refer to RGB and 3D data), it analyses the advantage of utilizing normalizing flows and enhances student-teacher networks by integrating a bijective normalizing flow as the teacher.
GNF \cite{2019_NeurIPS_GNF} introduces a reversible GNN model for graph generation and prediction tasks. In the supervised scenario, it passes messages using significantly less memory, enabling it to scale to larger graphs. In the unsupervised scenario, it integrates graph normalizing flows with an innovative graph auto-encoder to develop a generative model of graph structures.

\subsection{Knowledge Distillation}
Knowledge distillation involves transferring knowledge from a large and complex teacher model to a structurally simpler and smaller student model effectively \cite{2021_IJCV_survey, 2021_TPAMI_review}. T2-GNN \cite{2023_AAAI_T2-GNN} focuses on the challenge that when both features and structure are incomplete, the disparity between them caused by missing randomness intensifies their mutual interference, potentially leading to incorrect completions that adversely impact node representation. It independently designs feature-level and structure-level teacher models to offer specific guidance for the student model.
% , and it proposes a dual distillation mode to ensure the student model can learn better.
MuGSI \cite{2024_WWW_MuGSI} proposes a GNN-to-MLP distillation framework for graph classification task, and it enables efficient structural knowledge distillation at various granular levels (i.e., graph-level, subgraph-level, and node-level distillation).
GLocalKD \cite{2022_WSDM_GLocalKD} applies knowledge distillation into graph-level anomaly detection, it achieves a random distillation by training one GNN encoder to predict another GNN encoder whose network parameters are randomly initialized. From an architectural perspective, it also employs an asymmetric network structure.
GLADST \cite{2023_ADMA_GLADST} employs a knowledge distillation framework consisting of one teacher model and two student models to conduct graph-level anomaly detection. In this dual-students-teacher model, the teacher model, guided by a heuristic loss, is trained to enhance the divergence of graph representations. 

\section{Method}
%--------------------- Fig 2 framework ----------------------%
\begin{figure}[!ht]
    \vspace{-1em}
    \centering
    \includegraphics[width=\textwidth]{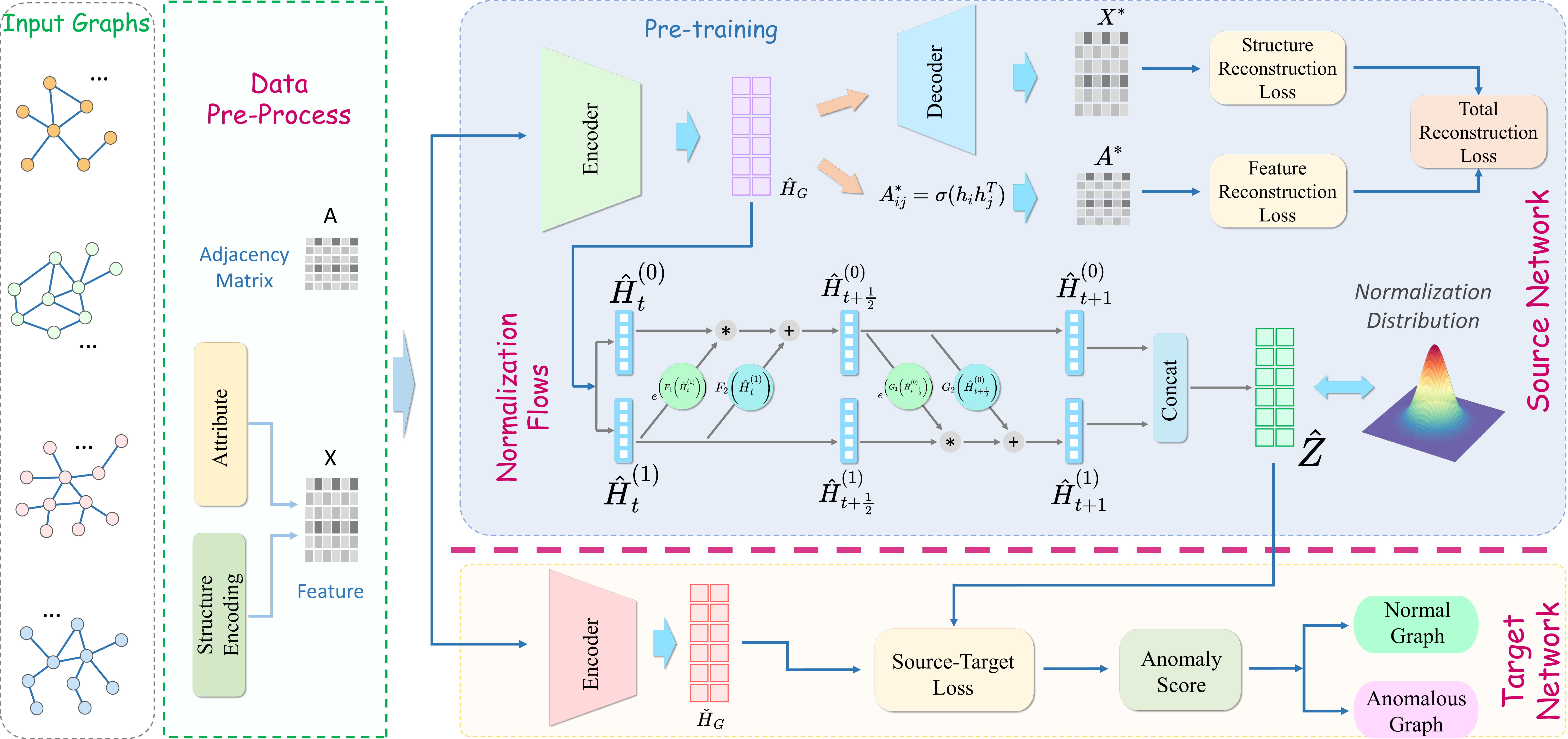}
    \caption{The overall framework of FANFOLD.} %{There are 4 stages in the \textcolor{darkbrown}{training phase}: ($I$) \textit{Data Pre-Processing}: when inputting a graph, we adopt a commonly used data augmentation approach, which not only makes good use of the node's attributes but also performs structural encoding to better capture and utilize structural information, forming the initial representation $X$ of the node. ($II$) \textit{Source network encoder Pre-Training}: we train the encoder through a reconstruction task (i.e., structure and feature reconstruction). ($III$) \textit{normalizing flows}: we use the well-pre-trained source network encoder to obtain embedding $h_1$ and normalizing flow (NF) to learn an embedding $z_1$ that approximately follows a normal distribution $N(0, I)$. ($IV$) \textit{Transfer the knowledge from source network to target network}: we let the embedding $h_2$ obtained by the target network encoder to approximate the embedding $z_1$ to learn the source network model. In the \textcolor{darkbrown}{inference phase}, we utilize the source-target loss (the difference between source network and target network) as the anomaly score to discriminate normal graphs and anomalous graphs.} 
    \label{fig: framework}
\end{figure}
In this section, we provide the details of FANFOLD. The framework is shown in Figure \ref{fig: framework}, which consists of four key stages: \textit{Data Pre-Processing}, \textit{Source network encoder Pre-Training}, \textit{Normalizing Flows} and \textit{Target network}.

\subsection{Preliminaries}
In this work, an undirected attributed graph is defined as $G = (\mathcal{V}, \mathcal{E}, \mathbf{X})$,
where $\mathcal{V}$ is the set of nodes and $\mathcal{E}$ represents the set of edges. The topology information of $G$ is represented by an adjacent matrix $\mathbf{A} \in \mathbb{R}^{n \times n}$, with $n$ being the number of nodes. $\mathbf{A}_{i, j} = 1$ when an edge exists between node $v_i$ and node $v_j$, otherwise $\mathbf{A}_{i, j} = 0$.
$\mathbf{X} \in \mathbb{R}^{n \times {d_{attr}}}$ denotes the feature matrix of nodes. Each row in $\mathbf{X}$ corresponds to the feature vector of a node, with $d_{attr}$ dimensions. The set of graphs is represented as $\mathcal{G} = \{G_1, G_2, ..., G_m \}$, where $m$ indicates the number of graphs in the set $\mathcal{G}$.
During the training phase, the model is exclusively trained on normal graphs. In the inference phase, FANFOLD aims to identify and distinguish the anomalous graphs from the normal ones given a graph set $\mathcal{G}$.

\subsection{Data Processing}
% In order to make full use of both attribute and topology information of graphs, we employ perturbation-free graph augmentation strategy \cite{2023_WSDM_GOOD-D, 2023_AAAI_FedStar} to get structure encoding $\mathbf{X}_{struc}$, which focuses on the topology of graphs. After we obtain the attribute and structure of graphs, for simplicity and efficiency, we directly concatenate them to form the initial feature representation 
For a graph, both node attribute and graph topology information are important for UAGD tasks. Thus, we first employ perturbation-free graph augmentation strategy \cite{2023_WSDM_GOOD-D, 2023_AAAI_FedStar} to get structure encoding $\mathbf{X}_{struc}$. Then, we concatenate structure encoding and attribute feature to form the initial feature representation by $\mathbf{X}_{init} = [\mathbf{X} ~||~ \mathbf{X}_{struc}]~$,
where $||$ indicates concatenation operation. Based on this processing, we can make use of both the node attribute and topology information of graphs. 
%For graphs without node attributes, the initial feature representation is obtained by $\mathbf{X}_{init} = [ \mathbf{X}_{struc}]~$.

% \begin{equation}
%     \mathbf{X}_{init} = [\mathbf{X} ~||~ \mathbf{X}_{struc}]~,
% \end{equation}

\subsection{Source Network Encoder Pre-Training}
In this section, we first design an encoder utilizing the Graph Neural Network (GNN) to map graphs into a structured continuous space. 
The encoder takes the$\{A\}$ and $\{X_{init}\}$ as input, and output nodes embeddings $\hat{H}_G =\{{h_i}\}_{i=1}^{n} \in \mathbb{R} ^{n\times d}$.
To preserve the complex structure characteristics,
we use the binary cross entropy loss function as Eq. 1 to approximate the adjacent matrix, which can also make connected nodes closer in embedding space.

% structure We use the following binary cross entropy loss function to make connected nodes closer in embedding space:

% To enable the source model to effectively capture the complex structure and distribution characteristics of graph data, in addition to data augmentation on the original feature vectors, we first utilize a Graph Neural Network (GNN) to preprocess the graph data, mapping it into a structured continuous space. A similar strategy has also been employed in previous works\cite{liu2019graph,garcia2021n}. The GNN encoder takes $G = (A, \mathbf{X})$ as input and output node embeddings $\hat{H}_{G}=\in \mathbb{R} ^{n\times d}$.  We use the following binary cross entropy loss function to make connected nodes closer in embedding space:

\begin{equation}
    \mathcal{L}_1= -\sum_{i = 1}^{N} \sum_{j = 1}^{\frac{N}{2}} A_{i j} \log \left({A}_{i j}^*\right)+\left(1-A_{i j}\right) \log \left(1-{A}_{i j}^*\right),
    \label{eq: l1}
\end{equation}
${A}_{i j}^*$ is the reconstructed ${A}_{i j}$, where ${A}_{i j}^*=\sigma (h_{i}h_{j}^T)$ and $\sigma (\cdot )$ denotes an activation function. 
Furthermore, we employ a decoder to reconstruct the original feature information. The simple Frobenius norm is used to design the loss function as Eq. 2.
\begin{equation}
     \mathcal{L}_{source}= (1- \alpha)~\mathcal{L}_1+\alpha~\|\mathbf{X}_{init}-\mathbf{X}^*\|_{F}^{2}~.
    \label{eq: source}
\end{equation}
$\mathbf{X}^*$ is the reconstructed $\mathbf{X}_{init}$, where $\mathbf{X}^*=  \varphi ^ {n} (\{{h_i}\})$. $\varphi ^ {n}(\cdot)$denotes n-layer perceptron, and $\alpha$ is a trade-off parameter. 
Based on Eq. 1 and Eq. 2, the encoder can effectively capture the complex structure characteristics as well as node feature information of graph data.

\subsection{Normalizing Flows}

The normalizing flows acts as the bijector between feature space and normalization latent space, which means the normalizing flows transform the feature of the input graphs from the original distribution to the normal distribution (i.e., the standard normalization distribution). We model $\hat{Z}_{T} \sim \prod_{i=1}^{N} \mathcal{N}\left(\mathbf{z}_{i} \mid 0, I\right)$ in our normalization latent space. Next, we will focus on the implementation process. To achieve reversibility, the node embedding $\hat{H}_G$ from the GNN is split into two components along the embedding dimension: $\hat{H}^{(0)}_t$ and $\hat{H}^{(1)}_t$. The process of normalizing flows can be described in detail as follows:
\begin{equation}
\begin{aligned}
& \hat{H}_{t+\frac{1}{2}}^{(0)}=\hat{H}_t^{(0)} \odot e^{\left(F_1\left(\hat{H}_t^{(1)}\right)\right)}+F_2\left(\hat{H}_t^{(1)}\right) \quad \hat{H}_{t+1}^{(0)}=\hat{H}_{t+\frac{1}{2}}^{(0)} \\
& \hat{H}_{t+\frac{1}{2}}^{(1)}=\hat{H}_t^{(1)} \quad \hat{H}_{t+1}^{(1)}=\hat{H}_{t+\frac{1}{2}}^{(1)} \odot e^{ \left(G_1\left(\hat{H}_{t+\frac{1}{2}}^{(0)}\right)\right)}+G_2\left(\hat{H}_{t+\frac{1}{2}}^{(0)}\right),\\
&
\end{aligned}
\label{eq: nff}
\end{equation}
where $\odot$ represents the element-wise product, $t$ is the step of message passing. $F$ and $G$ are sub-networks that can be considered as Message Passing Neural Networks (MPNNs). Finally, $\hat{Z}=\operatorname{concat}\left(\hat{H}_T^{(0)}, \hat{H}_T^{(1)}\right)$
is used as the output of the source network. $concat(\cdot)$ is the concatenation
of $\hat{H}_{T}^{(0)}$ and $\hat{H}_{T}^{(1)}$.
Inspired by the work of Liu et al.\cite{liu2019graph}, we use a change of variables to provide the rule for exact density transformation. We assume \( H_t \sim P(H_t) \), then the density in terms of \( P(H_{t-1}) \) is given by:
\begin{equation}
P(H_{t-1}) = \det\left|\frac{\partial H_{t-1}}{\partial H_t}\right| P(H_t).
\end{equation}
Thus, we have the following equation:
\begin{equation}
P(Z)=\operatorname{det}\left|\frac{\partial \hat{H}_{T}}{\partial \hat{H}_{0}}\right| P\left(\hat{H}_{T}\right)=P\left(\hat{H}_{T}\right) \prod_{t=1}^{T} \operatorname{det}\left|\frac{\partial \hat{H}_{t}}{\partial \hat{H}_{t-1}}\right|,
\end{equation}
where \( H_0 \) is the input node embedding. The Jacobian matrices are lower triangular, hence making density computations tractable, $det|\cdot|$ is Jacobian determinant.
We minimize the negative log likelihood with $P(\hat{H}_{T})$ as the normal distribution by the following equation:
% \begin{equation}
% p(H_{t-1})=p(H_t)\left|\operatorname{det} \frac{\partial H_{t-1}}{\partial H_{t}}\right|
% \end{equation}

\begin{equation}
\mathcal{L}_{nf} = \frac{\left\|\hat{H}_T\right\|_2^2}{2} - \sum_{t=1}^{T} \log \operatorname{det}\left|\frac{\partial \hat{H}_{t}}{\partial \hat{H}_{t-1}}\right|
\label{eq: nf}
\end{equation}

\subsection{Target Network}

In contrast to the source network, the target network employs a GIN architecture. Although they differ in structure, they maintain consistency in input and output. The target network similarly computes node embeddings $\hat{H}_G$ based on the input $G = (A, \mathbf{X})$. To facilitate the subsequent computation of distances between the outputs of the two networks, it is essential to ensure that the dimensions of $\check{H}_G$ in the target network match those of the source network. Following the common practice in the field, we aim for consistency in both node-level and graph-level representations. We obtain the graph representation through the READOUT function, which can involve operations such as averaging, maximum pooling, or minimum pooling. In our work, we opt for maximum pooling to derive the graph representation. This process can be described as follows:
\begin{equation}
    h_{G}=[\max_{i=1}^{n}h_{i,1},\max_{i=1}^{n}h_{i,2},...,\max_{i=1}^{n}h_{i,d}].
\label{eq: readout}
\end{equation}
The target network is trained by minimizing the distance between its outputs and those of the source:
\begin{equation}
\hat{\mathcal{L}}_{graph}=\frac1{m}\sum_{G\in\mathcal{G}}f_d\left(\check{h}_G,\hat{z}_G\right),\\\hat{\mathcal{L}}_{node}=\frac1{m}\sum_{G\in\mathcal{G}}\left(\frac1{n}\sum_{v_i\in\mathcal{V}_G}f_d\left(\check{h}_i,\hat{z}_i\right)\right),
\end{equation}
where $m$ is the size of graph set and $n$ is the size of graph. $f_{d}$ represents an arbitrary vector distance calculation method. We jointly minimize the $\hat{\mathcal{L}}_{graph}$ and  $\hat{\mathcal{L}}_{node}$ to learn normal graph pattern and feature distribution:
\begin{equation}
\hat{\mathcal{L}}_{target}=(1-\beta)\hat{\mathcal{L}}_{graph}+\beta\hat{\mathcal{L}}_{node},
\label{eq: target}
\end{equation} 
where $\beta$ is a tunable balancing factor to weigh the impact of $\hat{\mathcal{L}}_{graph}$ and  $\hat{\mathcal{L}}_{node}$. In the inference stage, we employ the same strategy to estimate the anomaly level of each graph. The score for each graph is determined by:
\begin{equation}
Score_G=f_d\left(h_G,\hat{h}_G\right)+\frac1{n}\sum_{v_i\in\mathcal{V}_G}f_d\left(h_i,\hat{h}_i\right),
\label{eq: score}
\end{equation}
The underlying idea behind this computation method is that, after sufficient training, the source network captures the characteristic and distribution patterns of normal graphs, and this information is also acquired by the target network. Consequently, when the input $G$ is normal, the output of the target network closely aligns with that of the source network, leading to a $score$ approaching $0$. However, when the input $G$ is abnormal, the target network has not encountered such cases during training, resulting in increased disparity between the outputs of the two networks and hence a target $score$ close to $1$.

%=========================== Experiments =============================%
\section{Experiments}
%%%%%%%%%%%%%%%%%%%%%%%%% tab 1 %%%%%%%%%%%%%%%%%%%%%%%%
\begin{table*}[t]
\caption{The statistics of datasets used in this work \cite{2020_arXiv_TuDataset}.}
\label{datasets}
\centering
\scalebox{0.5}{
    \renewcommand{\arraystretch}{2.0}
    \begin{tabular}{c|ccccccccccccccc}
    \toprule
    \textbf{Dataset}    & \textbf{PROTEINS\_full} & \textbf{ENZYMES} & \textbf{AIDS} & \textbf{DHFR} & \textbf{BZR} & \textbf{COX2} & \textbf{DD} & \textbf{NCI1} & \textbf{IMDB-B} & \textbf{REDDIT-B} & \textbf{COLLAB} & \textbf{HSE} & \textbf{MMP} & \textbf{p53} & \textbf{PPAR-gamma} \\ \hline
    \textbf{Graphs}     & 1113                    & 600              & 2000          & 467           & 405          & 467           & 1178        & 4110          & 1000            & 2000              & 5000            & 8417         & 7558         & 8903         & 8451                \\ \hline
    \textbf{Avg. Nodes} & 39.06                   & 32.63            & 15.69         & 42.43         & 35.75        & 41.22         & 284.32      & 29.87         & 19.77           & 429.63            & 74.49           & 16.89        & 17.62        & 17.92        & 17.38               \\ \hline
    \textbf{Avg. Edges} & 72.82                   & 62.14            & 16.20         & 44.54         & 38.36        & 43.45         & 715.66      & 32.30         & 96.53           & 497.75            & 2457.78         & 17.23        & 17.98        & 18.34        & 17.72               \\ 
    \bottomrule
    \end{tabular}
    }
\end{table*}

\subsection{Experimental Setup}
\textbf{Datasets.} We conduct comprehensive experiments on 15 datasets from the widely adopted benchmark (i.e., TUDataset \cite{2020_arXiv_TuDataset}) on UGAD, which involves different fields (small molecules, bioinformatics, and social networks). We follow the setting in \cite{2022_WSDM_GLocalKD, 2023_WSDM_GOOD-D} to achieve the definition of anomalous graphs, while the rest are viewed as normal data (normal graphs). Similar to \cite{2021_BigData_OCGIN, 2022_WSDM_GLocalKD, 2023_WSDM_GOOD-D}, only normal data are utilized in the training phase. More details of the datasets are presented in Table \ref{datasets}.

\textbf{Baselines.} We compare the proposed model with nine representative baselines. Under \textit{the non-end-to-end method}, we focus on two categories: (i) kernel + detector. The Weisfeiler-Lehman kernel (WL) \cite{2011_JMLR_WL_graph_Kernel} and propagation kernel (PK) \cite{2016_ML_PK_graph_Kernel} are used to obtain representations. Then, we apply one-class SVM (OCSVM) \cite{2001_JMLR_OCSVM} and isolation forest (iF) \cite{2008_ICDM_iF_graph_Kernel} to detect anomalies. This combination yields four baselines: PK-OCSVM, PK-iF, WL-OCSVM and WL-iF; (ii) GCL model + detector. We select two classic graph-level contrastive learning models, i.e., InfoGraph \cite{2020_ICLR_InfoGraph} and GraphCL \cite{2020_NeurIPS_GraphCL}, to obtain representations first. We then use iF as the detector to identify anomalies, resulting in InfoGraph-iF, GraphCL-iF. For \textit{the end-to-end method}, we select three representative models: OCGIN \cite{2021_BigData_OCGIN}, GLocalKD \cite{2022_WSDM_GLocalKD} and GOOD-D \cite{2023_WSDM_GOOD-D}.

\textbf{Metrics and Implementations.}
Following \cite{2022_WSDM_GLocalKD, 2023_WSDM_GOOD-D, 2022_ScientificReports_GLADC, 2023_ECMLPKDD_CVTGAD}, we utilize the widely adopted graph-level anomaly detection metric (i.e., the area under the receiver operating characteristic (AUC)) to evaluate model performance. A higher AUC value corresponds to better anomaly detection performance. In practice, we employ Pytorch \cite{2019_NeurIPS_PyTorch_Library} to implement FANFOLD.

%%%%%%%%%%%%%%%%%%%%%%%%%%% Table: overall performance %%%%%%%%%%%%%%%%%%%%%%%%%%%%%
\begin{table}[t]
\caption{The performance comparison in terms of AUC (in percent, mean value ± standard deviation). The best performance is highlighted in \textbf{\textcolor{darkblue}{bold}}, and the second-best performance is \textcolor{lightblue}{{\ul underlined}}. \dag: we report the result from \cite{2023_WSDM_GOOD-D}.}
\label{overall_performance}
\centering
\scalebox{0.53}{
    \renewcommand{\arraystretch}{1.9}
    \begin{tabular}{l|cccccccccc}
    \hline
    Method                 & \textbf{PK-OCSVM\dag} & \textbf{PK-iF\dag} & \textbf{WL-OCSVM\dag} & \textbf{WL-iF\dag} & \textbf{InfoGraph-iF\dag} & \textbf{GraphCL-iF\dag} & \textbf{OCGIN\dag}      & \textbf{GLocalKD\dag}   & \textbf{GOOD-D\dag}     & \textbf{FANFOLD}     \\ \hline
    \textbf{PROTEINS-full} & 50.49±4.92        & 60.70±2.55     & 51.35±4.35        & 61.36±2.54     & 57.47±3.03            & 60.18±2.53          & 70.89±2.44          & \textbf{\textcolor{darkblue}{77.30±5.15}} & \textcolor{lightblue}{\ul 71.97±3.86}          &71.10±1.30 \\
    \textbf{ENZYMES}       & 53.67±2.66        & 51.30±2.01     & 55.24±2.66        & 51.60±3.81     & 53.80±4.50            & 53.60±4.88          & 58.75±5.98          & \textcolor{lightblue}{\ul61.39±8.81}          & \textbf{\textcolor{darkblue}{63.90±3.69}}    & 59.19±2.73 \\
    \textbf{AIDS}          & 50.79±4.30        & 51.84±2.87     & 50.12±3.43        & 61.13±0.71     & 70.19±5.03            & 79.72±3.98          & 78.16±3.05          & 93.27±4.19          & \textcolor{lightblue}{{\ul 97.28±0.69}}    & \textbf{\textcolor{darkblue}{99.44±0.12}} \\
    \textbf{DHFR}          & 47.91±3.76        & 52.11±3.96     & 50.24±3.13        & 50.29±2.77     & 52.68±3.21            & 51.10±2.35          & 49.23±3.05          & 56.71±3.57          & \textcolor{lightblue}{{\ul 62.67±3.11}}    & \textbf{\textcolor{darkblue}{63.89±3.83}} \\
    \textbf{BZR}           & 46.85±5.31        & 55.32±6.18     & 50.56±5.87        & 52.46±3.30     & 63.31±8.52            & 60.24±5.37          & 65.91±1.47          & 69.42±7.78          & \textcolor{lightblue}{\ul {75.16±5.15}}    & \textbf{\textcolor{darkblue}{77.50±5.63}} \\
    \textbf{COX2}          & 50.27±7.91        & 50.05±2.06     & 49.86±7.43        & 50.27±0.34     & 53.36±8.86            & 52.01±3.17          & 53.58±5.05          & 59.37±12.67         & \textcolor{lightblue}{{\ul 62.65±8.14}}    & \textbf{\textcolor{darkblue}{64.28±3.85}} \\
    \textbf{DD}            & 48.30±3.98        & 71.32±2.41     & 47.99±4.09        & 70.31±1.09     & 55.80±1.77            & 59.32±3.92          & 72.27±1.83          & \textcolor{lightblue}{\ul {80.12±5.24}} & 73.25±3.19          & \textbf{\textcolor{darkblue}{82.33±0.84}}    \\
    \textbf{NCI1}          & 49.90±1.18        & 50.58±1.38     & 50.63±1.22        & 50.74±1.70     & 50.10±0.87            & 49.88±0.53          & \textbf{\textcolor{darkblue}{71.98±1.21}} & \textcolor{lightblue}{{\ul 68.48±2.39}}    & 61.12±2.21          & 62.00±4.76\\
    \textbf{IMDB-B}        & 50.75±3.10        & 50.80±3.17     & 54.08±5.19        & 50.20±0.40     & 56.50±3.58            & 56.50±4.90          & 60.19±8.90          & 52.09±3.41          & \textcolor{lightblue}{{\ul 65.88±0.75}}    & \textbf{\textcolor{darkblue}{68.94±1.46}} \\
    \textbf{REDDIT-B}      & 45.68±2.24        & 46.72±3.42     & 49.31±2.33        & 48.26±0.32     & 68.50±5.56            & 71.80±4.38          & 75.93±8.65          & 77.85±2.62          & \textbf{\textcolor{darkblue}{88.67±1.24}} & \textcolor{lightblue}{{\ul79.27±3.22}}    \\
    \textbf{COLLAB}        & 49.59±2.24        & 50.49±1.72     & 52.60±2.56        & 50.69±0.32     & 46.27±0.73            & 47.61±1.29          & 60.70±2.97          & 52.94±0.85          & \textbf{\textcolor{darkblue}{72.08±0.90} }  & \textcolor{lightblue}{\ul60.78±7.51}\\
    \textbf{HSE}           & 57.02±8.42        & 56.87±10.51    & 62.72±10.13       & 53.02±5.12     & 53.56±3.98            & 51.18±2.71          & 64.84±4.70          & 59.48±1.44          & \textcolor{lightblue}{\ul {69.65±2.14}}    & \textbf{\textcolor{darkblue}{69.79±5.53}} \\
    \textbf{MMP}           & 46.65±6.31        & 50.06±3.73     & 55.24±3.26        & 52.68±3.34     & 54.59±2.01            & 54.54±1.86          & \textbf{\textcolor{darkblue}{71.23±0.16}} & 67.84±0.59          & 70.51±1.56          &63.87±2.00   \\
    \textbf{p53}           & 46.74±4.88        & 50.69±2.02     & 54.59±4.46        & 50.85±2.16     & 52.66±1.95            & 53.29±2.32          & 58.50±0.37          & \textcolor{lightblue}{\ul {64.20±0.81}}    & 62.99±1.55          & \textbf{\textcolor{darkblue}{64.41±0.42}} \\
    \textbf{PPAR-gamma}    & 53.94±6.94        & 45.51±2.58     & 57.91±6.13        & 49.60±0.22     & 51.40±2.53            & 50.30±1.56          & \textcolor{lightblue}{\ul {71.19±4.28}} & 64.59±0.67          & 67.34±1.71          & \textbf{\textcolor{darkblue}{71.22±2.09}}    \\ \hline
    Avg.Rank               & 8.73              & 7.73           & 6.93              & 7.47           & 6.53                  & 6.93                & 3.67                & 3.13                & 2.13                & 1.73                \\ \hline
    \end{tabular}
}
\end{table}

\subsection{Overall Performance} 
The AUC results of FANFOLD and nine baselines w.r.t AUC and Avg.Rank on 15 datasets are reported in Table \ref{overall_performance}. As shown in Table \ref{overall_performance}, our proposed FANFOLD achieves first place on 9 datasets, secured second place on 2 datasets, and demonstrates competitive performance on the remaining datasets. In addition, FANFOLD attains the highest average rank among all comparative methods across 15 datasets. Based on our observations, graph kernel-based methods show the weakest performance among tested baselines due to their limited capacity to detect regular patterns and crucial graph information, which hinders their effectiveness with complex datasets and leads to subpar results. GCL-based methods exhibit a moderate level of performance, indicating the competitive potential of graph contrastive learning for UGAD tasks. To conclude, the competitive performance of our proposed model demonstrates the necessity and superiority of implementing asymmetric network with normalizing flows for graph-level anomaly detection. Such results also prove that FANFOLD is intrinsically capable of capturing the core characteristics of normal graphs, thereby achieving superior anomaly detection performance.

%Based on our observations, graph kernel-based methods demonstrate the lowest performance among the evaluated baselines. This underperformance can likely be attributed to their insufficient ability to capture regular patterns and essential information within the graph structures. These methods may struggle to identify and utilize key features necessary for accurate analysis, leading to their comparative ineffectiveness in handling complex datasets. Consequently, their limited capability to discern critical structural information results in suboptimal outcomes. %This suggests that they are still capable and effective in addressing the given problem. 

\subsection{Ablation Study}
%%%%%%%%%%%%%%%%%%% Figure: Ablation Study %%%%%%%%%%%%%%%%%%%%
% \begin{wrapfigure}[13]{r}{0.72\textwidth}
%     \centering
%     \includegraphics[width=0.6\textwidth]{figs/experiment/ablation.pdf}
%     \caption{Compare AUC results with different network architecture.}
%     \label{fig: ablation}
% \end{wrapfigure}
% \begin{wrapfigure}[18]{r}{0.66\textwidth}

\begin{figure}
    \centering
    \subfigure[AIDS]{\includegraphics[width=0.3\textwidth]{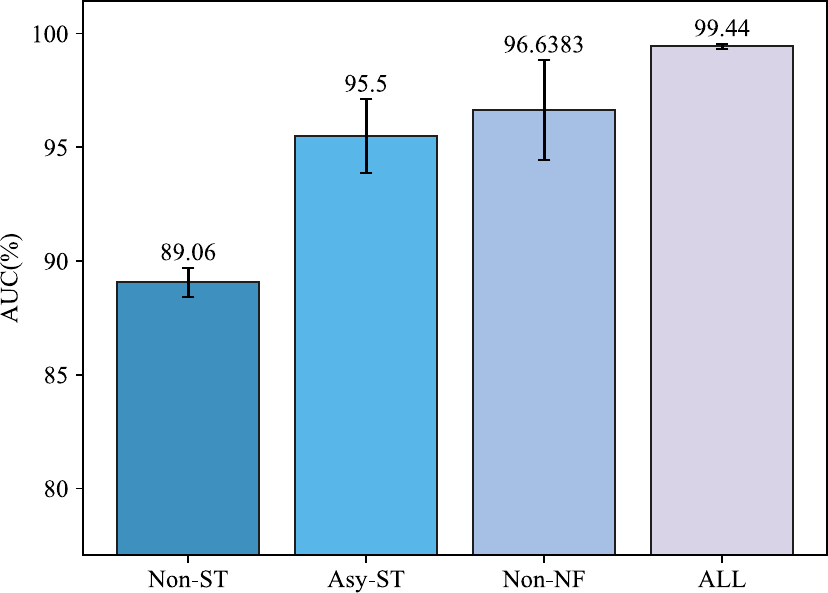}}
    \subfigure[DD]{\includegraphics[width=0.3\textwidth]{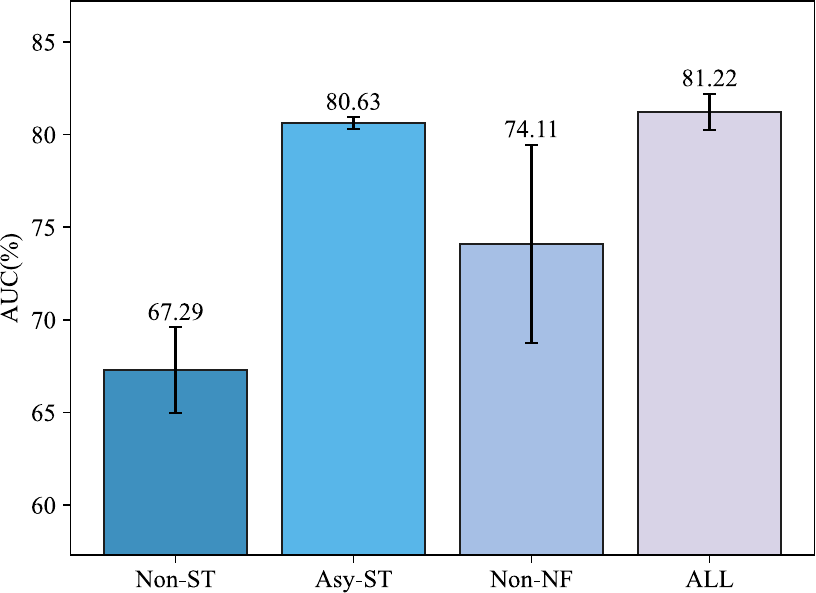}} 
    \subfigure[IMDB-BINARY]{\includegraphics[width=0.3\textwidth]{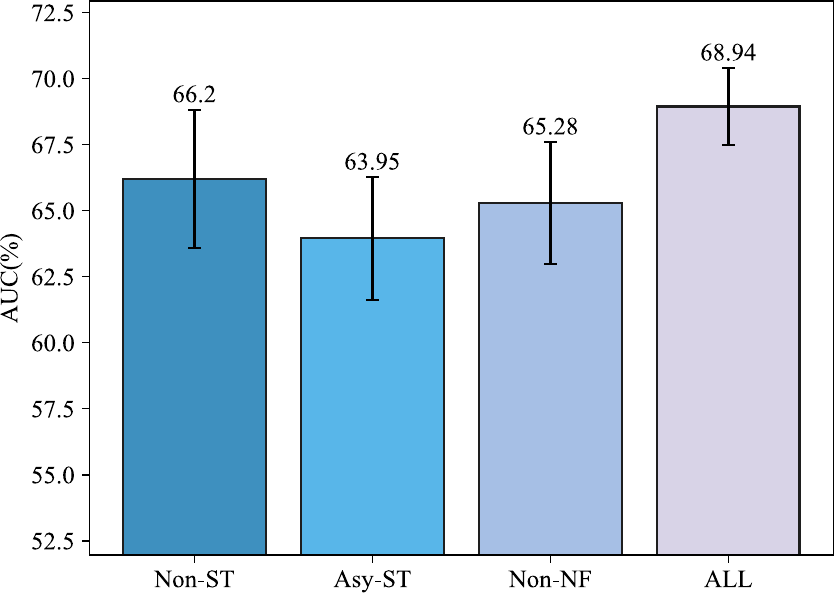}}    
    \subfigure[HSE]{\includegraphics[width=0.3\textwidth]{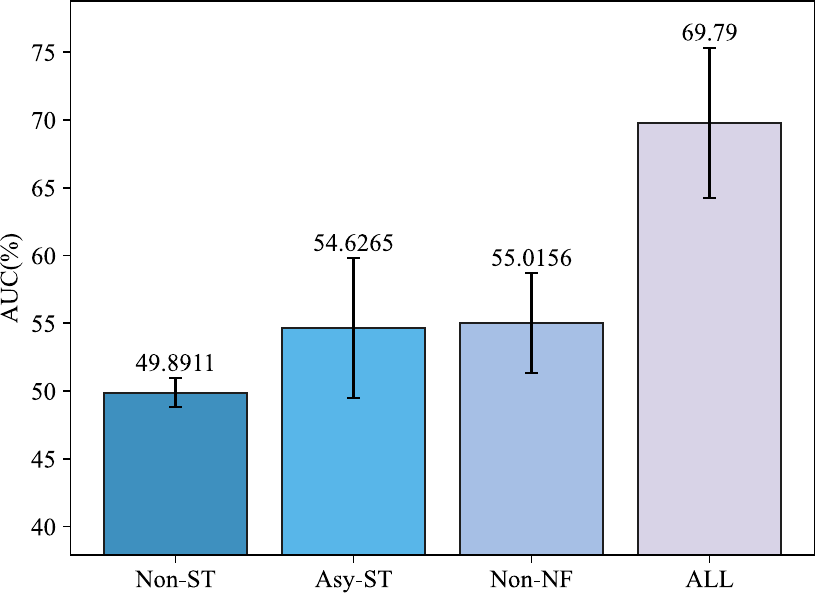}}
    \subfigure[PPAR-gamma]{\includegraphics[width=0.3\textwidth]{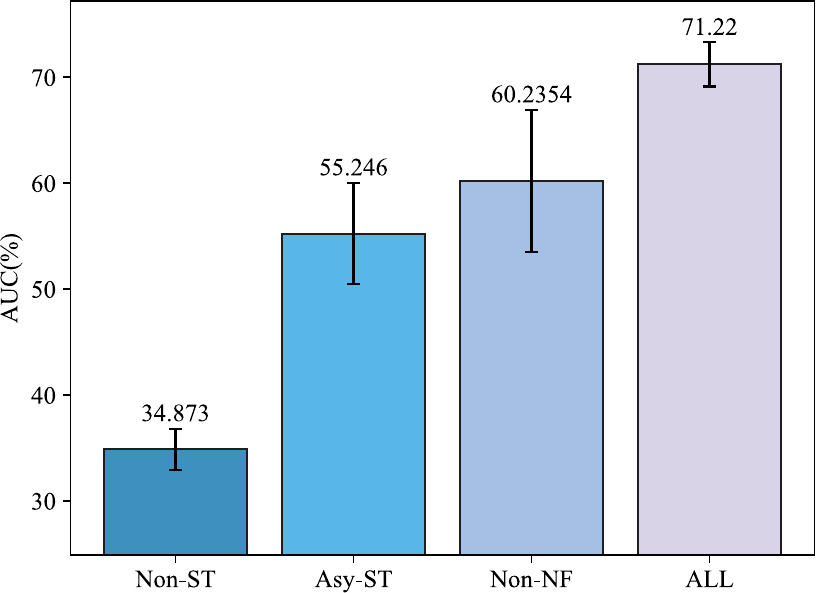}} 
    \subfigure[BZR]{\includegraphics[width=0.3\textwidth]{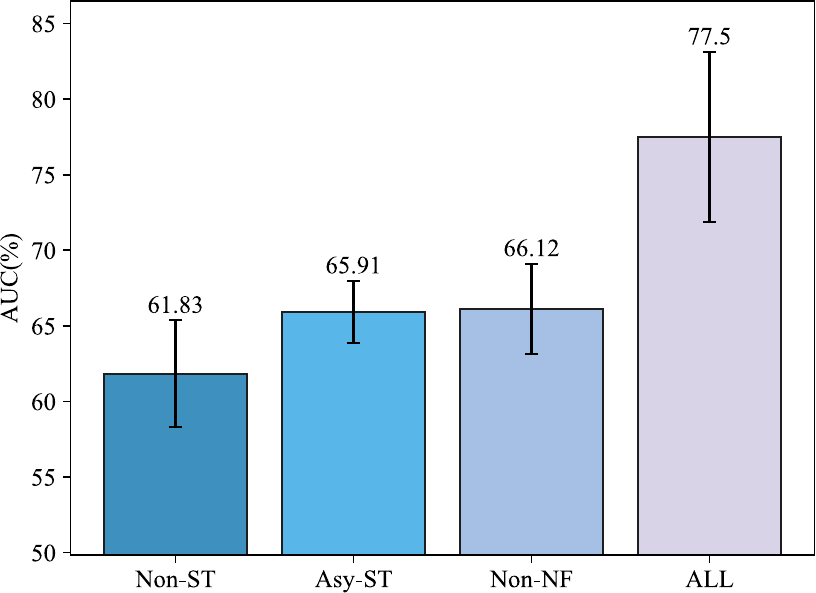}}
    \caption{AUC results of different network architectures.}
    \label{fig: ablation}
% \end{wrapfigure}
\end{figure}

In this study, we conduct a comprehensive ablation study to explore the importance of the target network and source network structure in FANFOLD.  The results are shown in Figure \ref{fig: ablation}. To this end, we design the following variant models for comparative analysis: 
(1) Non-ST: it only retains the source network and performs anomaly detection through reconstruction loss. Non-ST performs relatively poor, with its average AUC results dropping by approximately 15\% compared to other variants. This indicates that a single network has limitations in capturing complex features of the data and handling various data types.
(2) Asy-ST: it employs a symmetric source-target network. This variant shows significant performance improvement, indicating that constructing a symmetric ST network enables the model to capture the diversity and complexity of the data more effectively. 
(3) Non-NF: it adopts an asymmetric ST network structure without introducing normalizing flows. Non-NF achieves about a 2\% performance improvement on multiple datasets, demonstrating that an asymmetric structure can enhance the differences between the outputs of the source network and the target network, thereby improving model performance. 
To conclude, FANFOLD introduces normalizing flows into the source network, which results in about 10\% performance improvement on the BZR and HSE datasets, which proves that normalizing flows allows the model to capture the essential characteristics of normal graphs. 

%FANFOLD can more effectively capture the diversity and complexity of the data, resulting in significant performance improvements in anomaly detection tasks.

\subsection{Hyper-Parameter Analysis}
\begin{figure}
 \centering
    \subfigure[AIDS]{\includegraphics[width=0.32\textwidth]{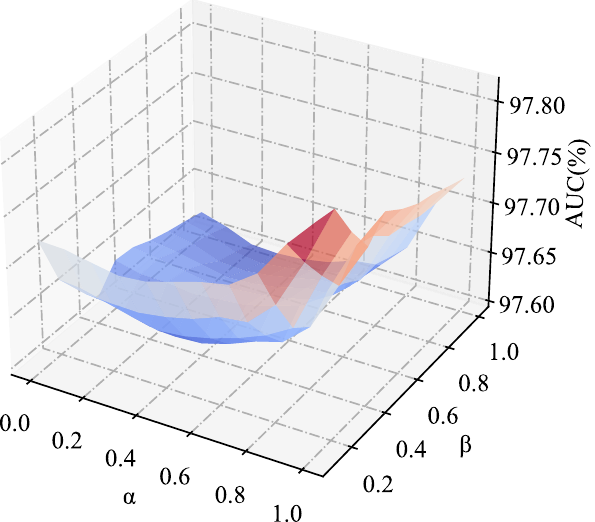}}
    \subfigure[HSE]{\includegraphics[width=0.32\textwidth]{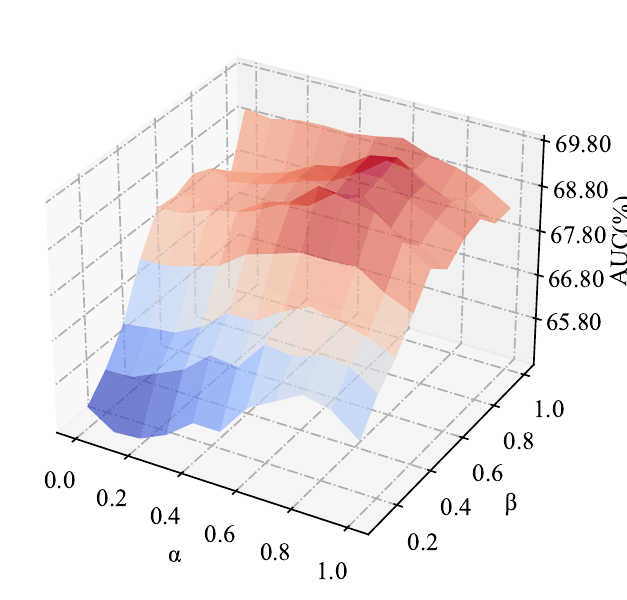}} 
    \subfigure[COX2]{\includegraphics[width=0.32\textwidth]{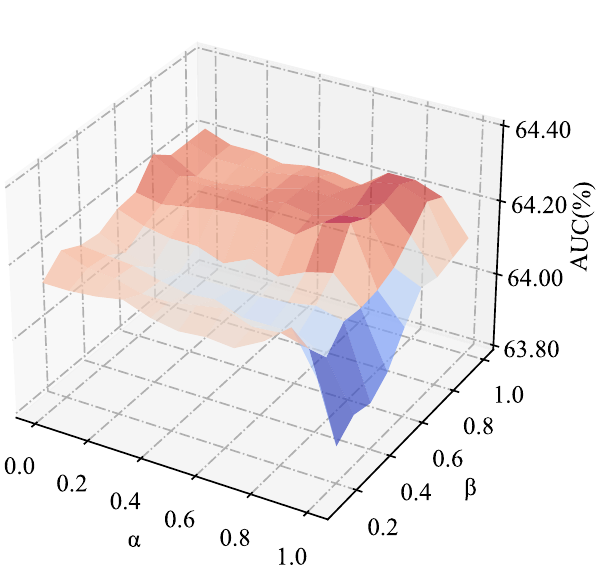}}
    \subfigure[IMDB-BINARY]{\includegraphics[width=0.32\textwidth]{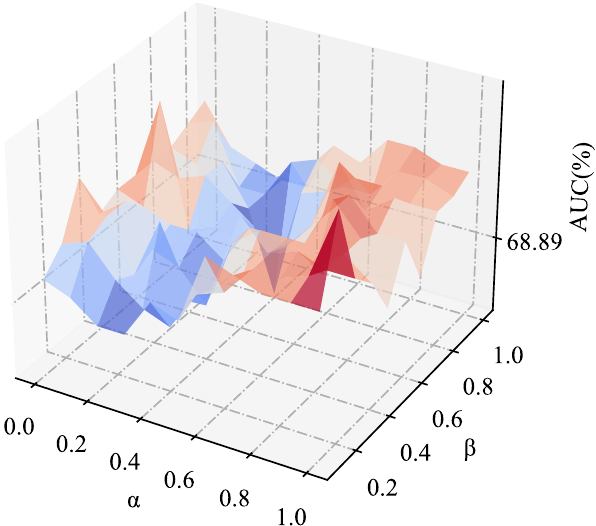}}
    \subfigure[DD]{\includegraphics[width=0.32\textwidth]{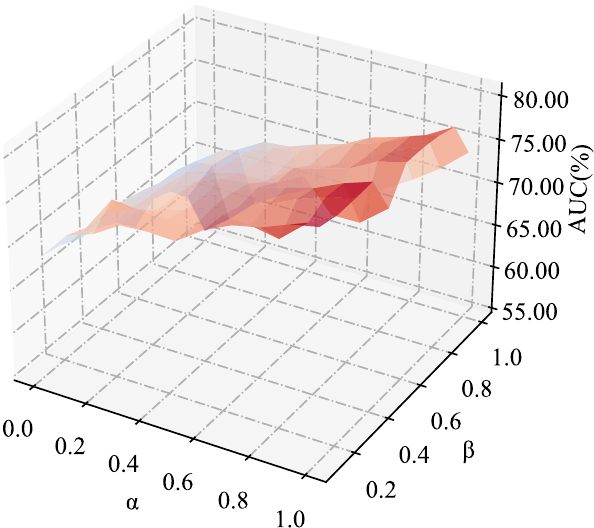}} 
    \subfigure[PPAR-gamma]{\includegraphics[width=0.32\textwidth]{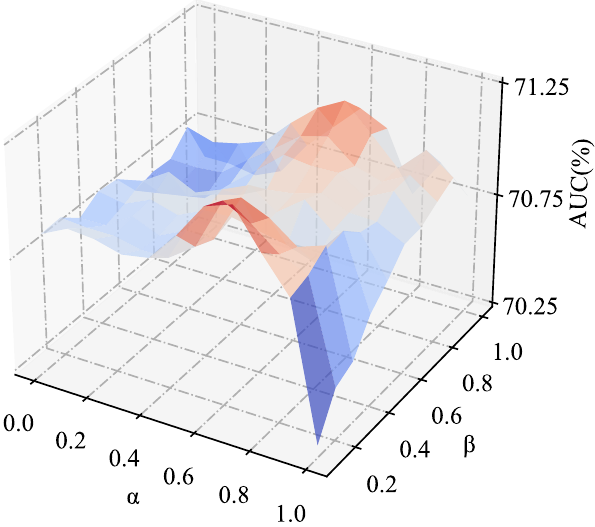}}
    \caption{Hyper-parameter Analysis: $\alpha$  and $\beta$.}
    \label{fig: 3d}
\end{figure}

\textbf{Balance Factors.} We explore the impact of the balance factors $\alpha$ and $\beta$ as defined in Equations \ref{eq: source} and \ref{eq: target}. $\alpha$ and $\beta$ are tuned in the range of $0\sim1$ respectively, and the results are illustrated in Figure \ref{fig: 3d}. As shown in Figure \ref{fig: 3d}, different datasets have varying sensitivity to $\alpha$ and $\beta$. For the AIDS and COX2 datasets, different balance factors have a minimal impact on the AUC value, showing relatively stable performance. In contrast, for the HSE, IMDB-BINARY, and PPAR-gamma datasets, parameter changes have a significant impact on AUC values, resulting in substantial fluctuations. For these datasets, adjusting parameters $\alpha$ and $\beta$ can optimize the AUC performance, where the highest AUC values occur when $\alpha$ is between 0.6 to 0.8 and $\beta$ is between 0.4 to 0.8.

\begin{wrapfigure}{r}{7cm}
    \centering
    \includegraphics[width=0.5\textwidth]{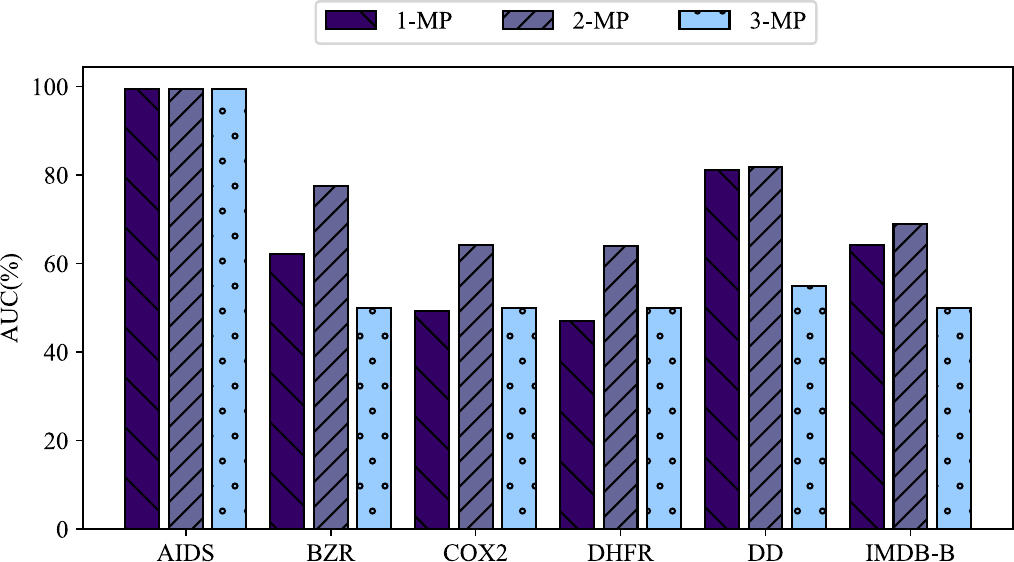}
    \caption{AUC w.r.t. MPs.}
    \label{fig: mp}
\end{wrapfigure}
\textbf{Message Passing Steps.} We study the impact of message passing (MP) steps in NF by varying the MP values in ${1, 2, 3}$. As shown in Figure \ref{fig: mp}, there is a significant improvement in AUC scores ranging from 10\% to 20\% when increasing the MP step from 1 to 2. This improvement may stem from the model's enhanced ability to propagate and aggregate information between nodes across two message-passing steps, effectively capturing the graph's global structure and local details. However, when increasing the MP from 2 to 3, there is a sharp performance decline across most datasets (e.g., except for the AIDS dataset). 
This could be attributed to unnecessary increases in model complexity due to additional message-passing steps. The model might overly focus on specific details/noises and neglect crucial global information, resulting in decreased performance. 
%Thus, the MP step is set to 2 in our experiments.

%%%%%%%%%%%%%%% Figure: Hyper-parameter Analysis %%%%%%%%%%%%%%%%%%%%
% \begin{wrapfigure}[20]{r}{0.66\textwidth}
% \end{wrapfigure}

\subsection{Visualization}

\begin{figure}[h]
\centering
\includegraphics[width=1\textwidth]{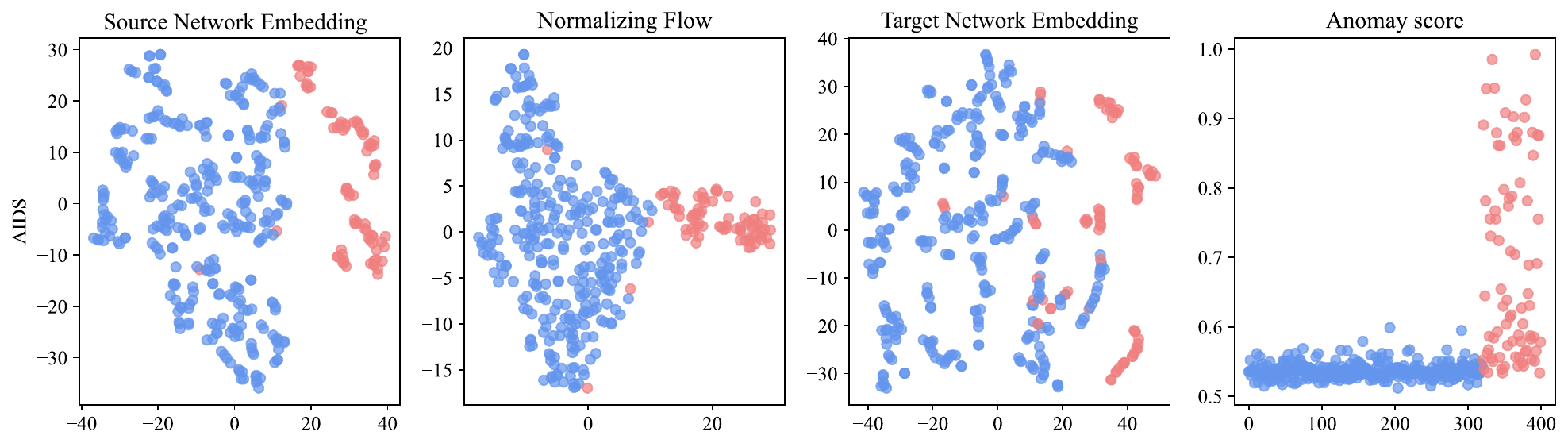}
\caption{Visualization Analysis. (\textcolor{blue}{$\bullet$} denotes normal, \textcolor{red}{$\bullet$} denotes anomaly.)}
\label{fig: vis}
% \end{wrapfigure}
\end{figure}

%%%%%%%%%%%%%%%% Figure: Visualization %%%%%%%%%%%%%%%%%%%
We utilize t-SNE for visualizing the graph embeddings learned by FANFOLD at different stages. The result is illustrated in Figure \ref{fig: vis}. After initial processing, the source network's embeddings show some separation between normal and anomalous graphs with noticeable overlap and broad distribution, indicating a partial but not substantial distinction between the sample types. Then, with normalizing flows, the separation between normal and anomalous samples becomes clearer and their distribution more concentrated. Also, it can be noticed that the distribution of anomalous samples becomes more scattered overall in target network embedding. This leads to a clearer distinction between the embedding of anomalous samples in the target network and the source network, ultimately achieving the goal of anomaly detection. In the anomalous evaluation stage, normal samples have lower anomaly scores, primarily concentrated around 0.5, while anomalous samples have higher anomaly scores, approaching 1. The experimental results demonstrate that the step-by-step processing through Source Network Embedding, Normalizing Flows, and Target Network Embedding progressively enhances the separation between samples, validating the effectiveness of the model in detecting anomalies.

%The final Anomaly Score clearly distinguishes normal and anomalous samples, validating the effectiveness of the model in handling and detecting anomalies.

\section{Conclusions and Limitations}
In this work, we propose a novel graph normalizing flows-driven asymmetric network for unsupervised graph-level anomaly detection, named FANFOLD. To the best of our knowledge, this is the first application of normalizing flows and asymmetric network in unsupervised graph-level anomaly detection. Specifically, we employ knowledge distillation techniques and apply normalizing flows to the source network. Such a method considers anomaly detection from the perspective of sample distribution and distribution density, capturing the essence of normal and anomalous graphs. In addition, the design of asymmetric network driven by normalizing flows helps to further distinguish the anomalous graphs from the normal ones. Extensive experiments on 15 datasets validate the superiority of FANFOLD. Considering limitations, we exercise strict control during the training phase to ensure that the training data does not contain any anomalous graphs. While it aligns with existing work, it diverges from real-world scenarios and incurs extra costs. %Additionally, there is a lack of an effective mechanism to ensure that the model's estimation of the normal graph distribution is as expected, which can undermine the model's performance. 
\newpage

\small
\bibliographystyle{unsrt}
\bibliography{FANFOLD}

\begin{thebibliography}{10}

\bibitem{2021_TKDE_Survey_GAD_maxiaoxiao}
Xiaoxiao Ma, Jia Wu, Shan Xue, Jian Yang, Chuan Zhou, Quan~Z Sheng, Hui Xiong, and Leman Akoglu.
\newblock A comprehensive survey on graph anomaly detection with deep learning.
\newblock {\em IEEE Transactions on Knowledge and Data Engineering}, 2021.

\bibitem{2022_WSDM_GLocalKD}
Rongrong Ma, Guansong Pang, Ling Chen, and Anton van~den Hengel.
\newblock Deep graph-level anomaly detection by glocal knowledge distillation.
\newblock In {\em Proceedings of the fifteenth ACM international conference on web search and data mining}, pages 704--714, 2022.

\bibitem{2023(2024)_NeurIPS_SIGNET}
Yixin Liu, Kaize Ding, Qinghua Lu, Fuyi Li, Leo~Yu Zhang, and Shirui Pan.
\newblock Towards self-interpretable graph-level anomaly detection.
\newblock {\em Advances in Neural Information Processing Systems}, 36, 2024.

\bibitem{2020_arXiv_TuDataset}
Christopher Morris, Nils~M Kriege, Franka Bause, Kristian Kersting, Petra Mutzel, and Marion Neumann.
\newblock Tudataset: A collection of benchmark datasets for learning with graphs.
\newblock {\em arXiv preprint arXiv:2007.08663}, 2020.

\bibitem{2020_CVPR_Uninformed_Students}
Paul Bergmann, Michael Fauser, David Sattlegger, and Carsten Steger.
\newblock Uninformed students: Student-teacher anomaly detection with discriminative latent embeddings.
\newblock In {\em Proceedings of the IEEE/CVF conference on computer vision and pattern recognition}, pages 4183--4192, 2020.

\bibitem{2022_IJCV_GCAD}
Paul Bergmann, Kilian Batzner, Michael Fauser, David Sattlegger, and Carsten Steger.
\newblock Beyond dents and scratches: Logical constraints in unsupervised anomaly detection and localization.
\newblock {\em International Journal of Computer Vision}, 130(4):947--969, 2022.

\bibitem{2021_CVPR_Anomaly_Detection}
Mariana-Iuliana Georgescu, Antonio Barbalau, Radu~Tudor Ionescu, Fahad~Shahbaz Khan, Marius Popescu, and Mubarak Shah.
\newblock Anomaly detection in video via self-supervised and multi-task learning.
\newblock In {\em Proceedings of the IEEE/CVF conference on computer vision and pattern recognition}, pages 12742--12752, 2021.

\bibitem{2021_arXiv_Student_Teacher_feature_pyramid}
Guodong Wang, Shumin Han, Errui Ding, and Di~Huang.
\newblock Student-teacher feature pyramid matching for anomaly detection.
\newblock {\em arXiv preprint arXiv:2103.04257}, 2021.

\bibitem{2021_Entropy_DTSNE}
Qinfeng Xiao, Jing Wang, Youfang Lin, Wenbo Gongsa, Ganghui Hu, Menggang Li, and Fang Wang.
\newblock Unsupervised anomaly detection with distillated teacher-student network ensemble.
\newblock {\em Entropy}, 23(2):201, 2021.

\bibitem{2023_WSDM_GOOD-D}
Yixin Liu, Kaize Ding, Huan Liu, and Shirui Pan.
\newblock Good-d: On unsupervised graph out-of-distribution detection.
\newblock In {\em Proceedings of the Sixteenth ACM International Conference on Web Search and Data Mining}, pages 339--347, 2023.

\bibitem{2023_DASFAA_TUAF}
Zhenyang Yu, Xinye Wang, Bingzhe Zhang, Zhaohang Luo, and Lei Duan.
\newblock Tuaf: Triple-unit-based graph-level anomaly detection with adaptive fusion readout.
\newblock In {\em International Conference on Database Systems for Advanced Applications}, pages 415--430. Springer, 2023.

\bibitem{2023_ECMLPKDD_CVTGAD}
Jindong Li, Qianli Xing, Qi~Wang, and Yi~Chang.
\newblock Cvtgad: Simplified transformer with cross-view attention for unsupervised graph-level anomaly detection.
\newblock In {\em Joint European Conference on Machine Learning and Knowledge Discovery in Databases}, pages 185--200. Springer, 2023.

\bibitem{2023_ECMLPKDD_HimNet}
Chaoxi Niu, Guansong Pang, and Ling Chen.
\newblock Graph-level anomaly detection via hierarchical memory networks.
\newblock In {\em Joint European Conference on Machine Learning and Knowledge Discovery in Databases}, pages 201--218. Springer, 2023.

\bibitem{2022_ScientificReports_GLADC}
Xuexiong Luo, Jia Wu, Jian Yang, Shan Xue, Hao Peng, Chuan Zhou, Hongyang Chen, Zhao Li, and Quan~Z Sheng.
\newblock Deep graph level anomaly detection with contrastive learning.
\newblock {\em Scientific Reports}, 12(1):19867, 2022.

\bibitem{2023_ADMA_GLADST}
Fu~Lin, Xuexiong Luo, Jia Wu, Jian Yang, Shan Xue, Zitong Wang, and Haonan Gong.
\newblock Discriminative graph-level anomaly detection via dual-students-teacher model.
\newblock In {\em International Conference on Advanced Data Mining and Applications}, pages 261--276. Springer, 2023.

\bibitem{2022_ICLR_GANF}
Enyan Dai and Jie Chen.
\newblock Graph-augmented normalizing flows for anomaly detection of multiple time series.
\newblock In {\em International Conference on Learning Representations}, 2022.

\bibitem{2021_BigData_OCGIN}
L~Zhao and L~Akoglu.
\newblock On using classification datasets to evaluate graph outlier detection: Peculiar observations and new insights.
\newblock {\em Big Data}, 11(3):151--180, 2021.

\bibitem{2019_ICLR_GIN}
Keyulu Xu, Weihua Hu, Jure Leskovec, and Stefanie Jegelka.
\newblock How powerful are graph neural networks?
\newblock {\em arXiv preprint arXiv:1810.00826}, 2018.

\bibitem{2020_TPAMI_Survey_Normalizing_Flows}
Ivan Kobyzev, Simon~JD Prince, and Marcus~A Brubaker.
\newblock Normalizing flows: An introduction and review of current methods.
\newblock {\em IEEE transactions on pattern analysis and machine intelligence}, 43(11):3964--3979, 2020.

\bibitem{2017_ICLR_RealNVP}
Laurent Dinh, Jascha Sohl-Dickstein, and Samy Bengio.
\newblock Density estimation using real nvp.
\newblock In {\em International Conference on Learning Representations}, 2017.

\bibitem{2021_WACV_DifferNet}
Marco Rudolph, Bastian Wandt, and Bodo Rosenhahn.
\newblock Same same but differnet: Semi-supervised defect detection with normalizing flows.
\newblock In {\em Proceedings of the IEEE/CVF winter conference on applications of computer vision}, pages 1907--1916, 2021.

\bibitem{2023_WACV_AST}
Marco Rudolph, Tom Wehrbein, Bodo Rosenhahn, and Bastian Wandt.
\newblock Asymmetric student-teacher networks for industrial anomaly detection.
\newblock In {\em Proceedings of the IEEE/CVF winter conference on applications of computer vision}, pages 2592--2602, 2023.

\bibitem{2019_NeurIPS_GNF}
Jenny Liu, Aviral Kumar, Jimmy Ba, Jamie Kiros, and Kevin Swersky.
\newblock Graph normalizing flows.
\newblock {\em Advances in Neural Information Processing Systems}, 32, 2019.

\bibitem{2021_IJCV_survey}
Jianping Gou, Baosheng Yu, Stephen~J Maybank, and Dacheng Tao.
\newblock Knowledge distillation: A survey.
\newblock {\em International Journal of Computer Vision}, 129(6):1789--1819, 2021.

\bibitem{2021_TPAMI_review}
Lin Wang and Kuk-Jin Yoon.
\newblock Knowledge distillation and student-teacher learning for visual intelligence: A review and new outlooks.
\newblock {\em IEEE transactions on pattern analysis and machine intelligence}, 44(6):3048--3068, 2021.

\bibitem{2023_AAAI_T2-GNN}
Cuiying Huo, Di~Jin, Yawen Li, Dongxiao He, Yu-Bin Yang, and Lingfei Wu.
\newblock T2-gnn: Graph neural networks for graphs with incomplete features and structure via teacher-student distillation.
\newblock In {\em Proceedings of the AAAI Conference on Artificial Intelligence}, volume~37, pages 4339--4346, 2023.

\bibitem{2024_WWW_MuGSI}
Tianjun Yao, Jiaqi Sun, Defu Cao, Kun Zhang, and Guangyi Chen.
\newblock Mugsi: Distilling gnns with multi-granularity structural information for graph classification.
\newblock In {\em Proceedings of the ACM on Web Conference 2024}, pages 709--720, 2024.

\bibitem{2023_AAAI_FedStar}
Yue Tan, Yixin Liu, Guodong Long, Jing Jiang, Qinghua Lu, and Chengqi Zhang.
\newblock Federated learning on non-iid graphs via structural knowledge sharing.
\newblock In {\em Proceedings of the AAAI conference on artificial intelligence}, 2023.

\bibitem{liu2019graph}
Jenny Liu, Aviral Kumar, Jimmy Ba, Jamie Kiros, and Kevin Swersky.
\newblock Graph normalizing flows.
\newblock {\em Advances in Neural Information Processing Systems}, 32, 2019.

\bibitem{2011_JMLR_WL_graph_Kernel}
Nino Shervashidze, Pascal Schweitzer, Erik~Jan Van~Leeuwen, Kurt Mehlhorn, and Karsten~M Borgwardt.
\newblock Weisfeiler-lehman graph kernels.
\newblock {\em Journal of Machine Learning Research}, 12(9), 2011.

\bibitem{2016_ML_PK_graph_Kernel}
Marion Neumann, Roman Garnett, Christian Bauckhage, and Kristian Kersting.
\newblock Propagation kernels: efficient graph kernels from propagated information.
\newblock {\em Machine Learning}, 102:209--245, 2016.

\bibitem{2001_JMLR_OCSVM}
Larry~M Manevitz and Malik Yousef.
\newblock One-class svms for document classification.
\newblock {\em Journal of machine Learning research}, 2(Dec):139--154, 2001.

\bibitem{2008_ICDM_iF_graph_Kernel}
Fei~Tony Liu, Kai~Ming Ting, and Zhi-Hua Zhou.
\newblock Isolation forest.
\newblock In {\em 2008 eighth ieee international conference on data mining}, pages 413--422. IEEE, 2008.

\bibitem{2020_ICLR_InfoGraph}
Fan-Yun Sun, Jordon Hoffman, Vikas Verma, and Jian Tang.
\newblock Infograph: Unsupervised and semi-supervised graph-level representation learning via mutual information maximization.
\newblock In {\em International Conference on Learning Representations}, 2020.

\bibitem{2020_NeurIPS_GraphCL}
Yuning You, Tianlong Chen, Yongduo Sui, Ting Chen, Zhangyang Wang, and Yang Shen.
\newblock Graph contrastive learning with augmentations.
\newblock {\em Advances in neural information processing systems}, 33:5812--5823, 2020.

\bibitem{2019_NeurIPS_PyTorch_Library}
Adam Paszke, Sam Gross, Francisco Massa, Adam Lerer, James Bradbury, Gregory Chanan, Trevor Killeen, Zeming Lin, Natalia Gimelshein, Luca Antiga, et~al.
\newblock Pytorch: An imperative style, high-performance deep learning library.
\newblock {\em Advances in neural information processing systems}, 32, 2019.

\end{thebibliography}

% \begin{thebibliography}{1}

% \bibitem{2023_WSDM_GOOD-D}
% Yixin Liu, Kaize Ding, Huan Liu, and Shirui Pan.
% \newblock Good-d: On unsupervised graph out-of-distribution detection.
% \newblock In {\em Proceedings of the Sixteenth ACM International Conference on Web Search and Data Mining}, pages 339--347, 2023.

% \end{thebibliography}

\newpage
% \section*{Appendix}
\appendix

\section{Pseudo Code}
\begin{algorithm}[H]
\SetKwInOut{Input}{Input}
\SetKwInOut{Output}{Output}
\SetKwInOut{Initialize}{Initialize}
\SetKwInOut{Parameter}{Parm}

\Input{Graph set:  $\mathcal{G} = \{G_1, G_2, ..., G_m \}$;}
\Output{The anomaly scores for each graph $Score_G$;}
\Initialize{The trainable parameters \(\hat{\Theta}\) and \(\tilde{\Theta}\) for the Source Network, and \(\check{\Theta}\) for the Target Network;}

\textbf{Training Phase} 

\For{$i = 1$ to $s\_epochs$}{
    Compute node embeddings $\hat{H}_G$ through GNN;\\
    Perform a gradient descent step on Eq. \ref{eq: source} w.r.t. $\hat{\Theta}$;
    }

\For{$i = 1$ to $n\_epochs$}{
    with fixed $\hat{\Theta}$:\\
        \hspace{0.5cm}Compute node embeddings $\hat{H}_G$\\
    Transform the node embeddings $\hat{H}_G$ to a latent vector $Z_G$ through Eq. \ref{eq: nff};\\
    Minimizing the loss Eq. \ref{eq: nf};
    }
    
\For{$i = 1$ to $t\_epochs$}{
    Compute node embeddings $\check{H}_G$;\\
    with fixed $\hat{\Theta},\tilde{\Theta}$:\\
        \hspace{0.5cm} Source-Network calculates  $\hat{Z}_G$;\\
    Compute graph representations  $\check{h}_G$ and  $\hat{z}_G$ through Eq. \ref{eq: readout};\\
    Perform a gradient descent step on Eq. \ref{eq: target} w.r.t. $\check{\Theta}$;
    }

\textbf{Inference Phase}

\For{$G_{i}$ in Graph set $G$}{
    Caculate anomaly scores via Eq.\ref{eq: score};
    }
    
\caption{FANFOLD}
\label{alg:algorithm1}
\end{algorithm}

\section{Supplement of Experiments}

\begin{figure}[h!]
    \centering
    \includegraphics[width=0.6\textwidth]{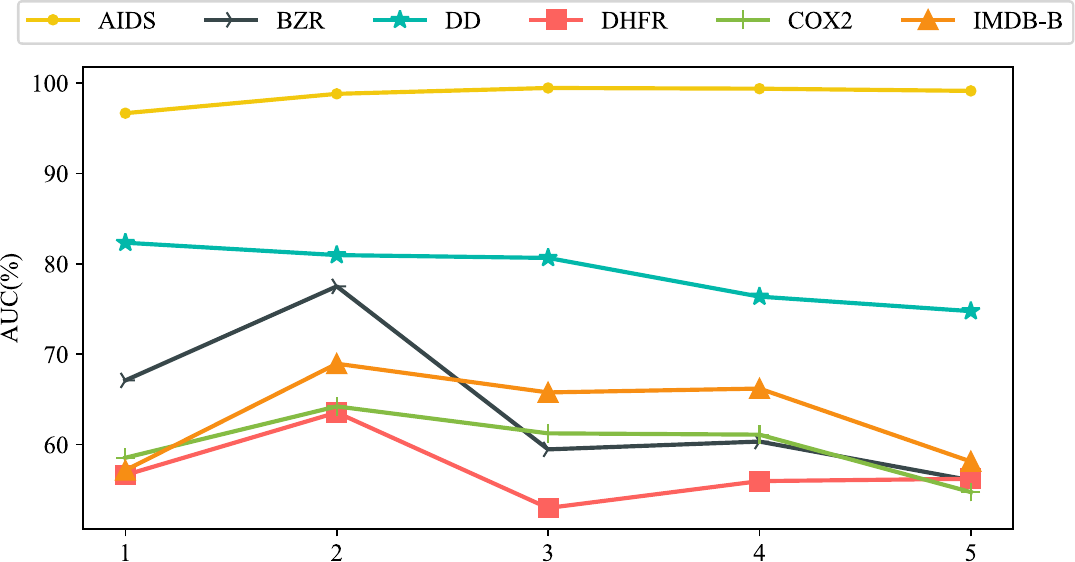}
    \caption{AUC w.r.t. GCN layer}
\label{fig: layer}
\end{figure}

\textbf{Number of GCN Layers.} 
We analyze the impact of GCN layers $k$ in the source network, where $k \in {1, 2, 3, 4, 5}$. The results are shown in Figure \ref{fig: layer}. Overall, most datasets achieve optimal results when the number of layers $k$ is 2. Beyond this, model performance either plateaus or gradually declines. Deeper GCNs do not improve performance but are more computationally costly.

\begin{figure}[ht]
    \centering
    \setcounter{subfigure}{0}
    \subfigure[AIDS]{\includegraphics[width=0.32\textwidth]{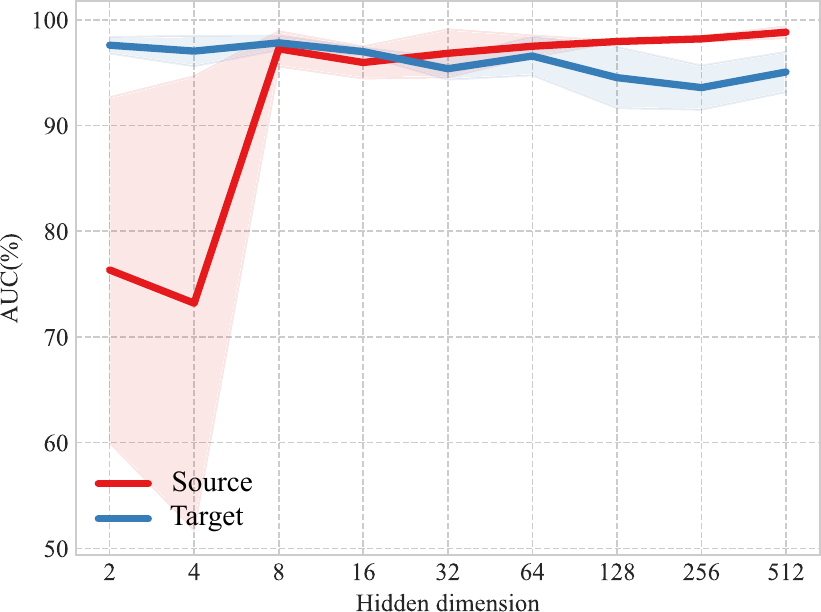}}
    \subfigure[BZR]{\includegraphics[width=0.32\textwidth]{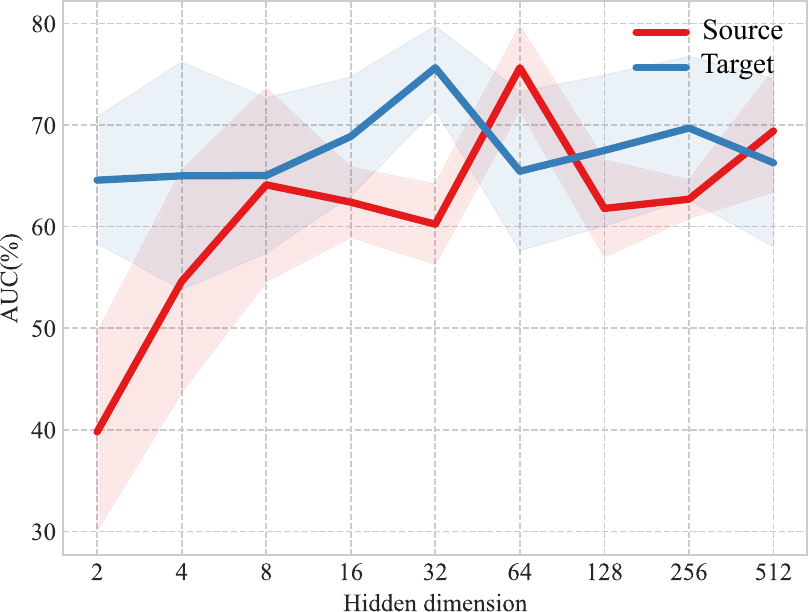}} 
    \subfigure[DD]{\includegraphics[width=0.32\textwidth]{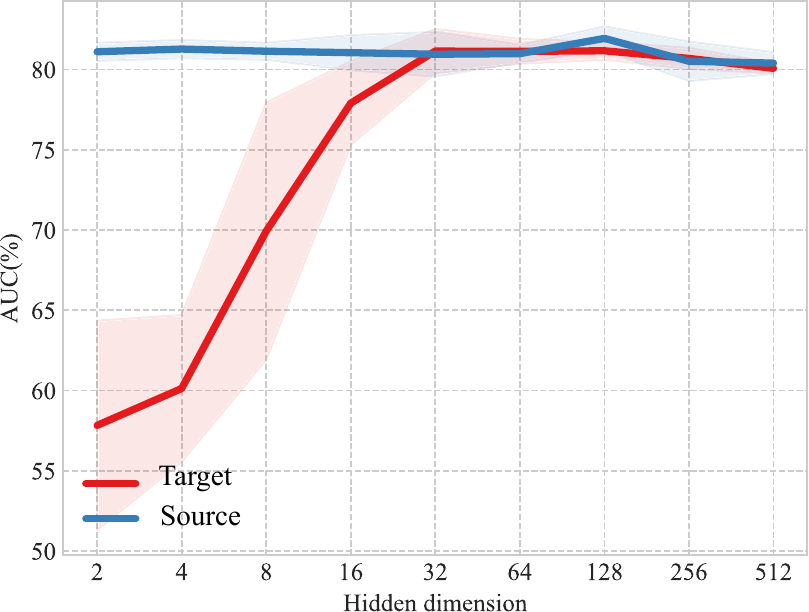}}
    \subfigure[COX2]{\includegraphics[width=0.32\textwidth]{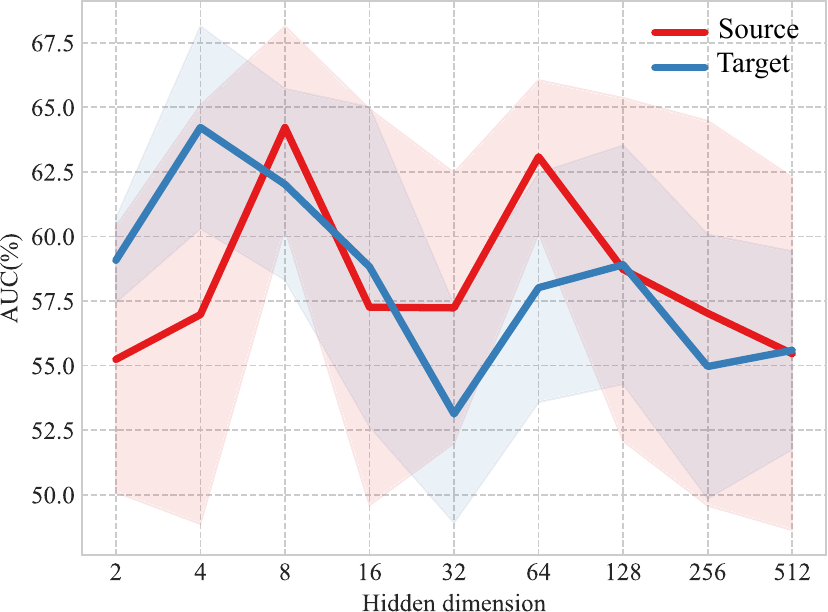}}
    \subfigure[IMDB-B]{\includegraphics[width=0.32\textwidth]{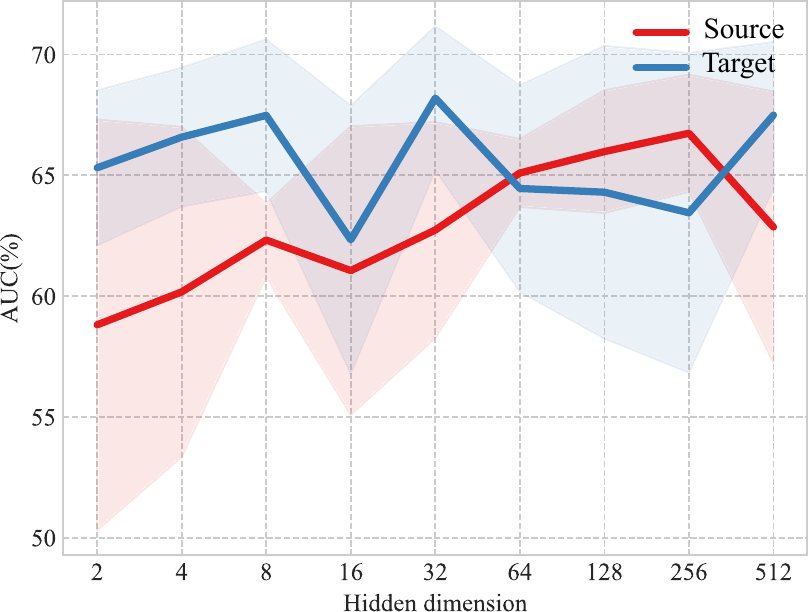}} 
    \subfigure[DHFR]{\includegraphics[width=0.32\textwidth]{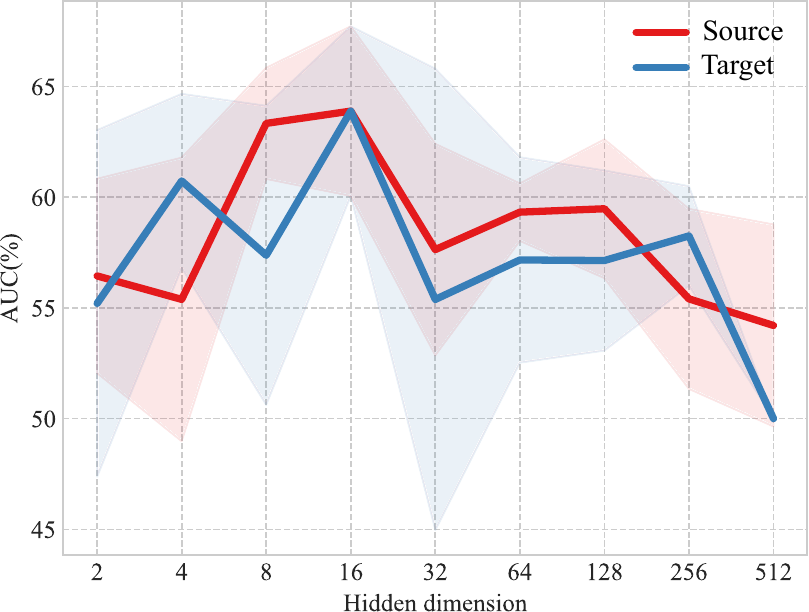}}
    \caption{Hyper-parameter Analysis: Hidden dimension}
    \label{fig: dim}
% \end{wrapfigure}
\end{figure}

\textbf{Dimensionality Size.}
To analyze the impact of dimensionality size of the source network and target network on model performance, we alter the hidden dimension from 2 to 512.  The AUC result on six dataset is shown in Figure \ref{fig: dim}. For a portion of the dataset, such as AIDS and DD, FANFOLD's AUC rapidly increases with the source network to highest value at hidden dimension $4\sim32$, then remains stable. At the same time, for these datasets, the model is relatively insensitive to changes in the dimensions of the target network, achieving optimal results around 8, 32, and 128. 
%%%%%%%%%%%%%%%% Figure: Hyper-parameter Analysis: Hidden dimension %%%%%%%%%%%%%%%%%%%
% \begin{wrapfigure}[18]{r}{0.66\textwidth}
In the datasets COX2, IMDB-BINARY, and DHFR, the model is more sensitive to the hidden layer dimensions of the source network and Tthe target network.
The overall trend shows that as the dimensions increase, the AUC fluctuates and rises until it reaches a peak, after which performance begins to decline. In most datasets, the source and target networks exhibit the highest AUC at specific dimensions, although the optimal dimensions vary. Additionally, different datasets have varying sensitivities to hidden dimensions, and selecting the appropriate hidden dimensions can significantly enhance the model's AUC performance.

% \textbf{Message passing steps.}
% We study the sensitivity of the FANFOLD model to the number of message passing (MP) steps in NF by varying the MP values as ${1, 2, 3}$. As shown in Figure \ref{fig: mp}, different datasets experience a significant improvement in AUC scores ranging from 10\% to 20\% when increasing the MP from 1 to 2. This improvement may stem from the model's enhanced ability to propagate and aggregate information between nodes across two message passing steps, thus better capturing both global structure and local features of the graph. However, when increasing the MP from 2 to 3, except for the AIDS dataset, there is a sharp decline in performance across other datasets. This could be attributed to unnecessary increases in model complexity due to the additional message passing steps, leading to overfitting or instability. With each additional message passing step, the model might overly focus on specific details or noise, neglecting more crucial global information, resulting in decreased performance across most datasets. Thus, overall, setting MP to 2 yields the best performance for most datasets as it improves performance without introducing overfitting or instability issues.
\subsection{Licenses}
\begin{itemize}
    \item \textbf{PyTorch:} \href{https://github.com/pytorch/pytorch/blob/main/LICENSE}{BSD-style}
    \item \textbf{Torchvision:} \href{https://github.com/pytorch/vision/blob/main/LICENSE}{BSD 3-Clause}
    \item \textbf{graph-generation:} \href{https://github.com/pytorch/vision/blob/main/LICENSE}{MIT}
    \item \textbf{G-OOD-D:} \href{https://github.com/yixinliu233/G-OOD-D/blob/main/LICENSE.md}{MIT}
\end{itemize}
%%%%%%%%%%%%%%%%%%%%%%%%%%%%%%%%%%%%%%%%%%%%%%%%%%%%%%%%%%%%

\newpage

\end{document}